\pdfoutput=1
\documentclass[sigconf, language=english]{acmart}

\acmConference[ICSE 2024]{46th International Conference on Software Engineering}{April 2024}{Lisbon, Portugal}
\usepackage{graphicx} 
\usepackage{pgfplots}
\pgfplotsset{width=5cm,compat=1.9}
\usepackage{color}
\definecolor{blue}{rgb}{0,0,1}
\newcommand{\revise}[1]{\textcolor{black}{#1}}

\copyrightyear{2024} 
\acmYear{2024} 
\setcopyright{rightsretained} 
\acmConference[ExEn '24 ]{2024 Workshop on Explainability Engineering}{April 20, 2024}{Lisbon, Portugal}
\acmBooktitle{2024 Workshop on Explainability Engineering (ExEn '24 ), April 20, 2024, Lisbon, Portugal}\acmDOI{10.1145/3648505.3648510}
\acmISBN{979-8-4007-0596-0/24/04}

\title{Why Reinforcement Learning in Energy Systems Needs Explanations}

\author{Hallah Shahid Butt}
\email{hallah.butt@kit.edu}
\affiliation{%
  \institution{Karlsruhe Institute of Technology}
  \city{Karlsruhe}
  \country{Germany}
}

\author{Benjamin Schäfer}
\email{benjamin.schaefer@kit.edu}
\affiliation{%
  \institution{Karlsruhe Institute of Technology}
  \city{Karlsruhe}
  \country{Germany}
}

\date{December 2023}

\begin{document}

\begin{abstract}
With economic development, the complexity of infrastructure has increased drastically. Similarly, with the shift from fossil fuels to renewable sources of energy, there is a dire need for such systems that not only predict and forecast with accuracy but also help in understanding the process of predictions. Artificial intelligence and machine learning techniques have helped in finding out well-performing solutions to different problems in the energy sector. However, the usage of state-of-the-art techniques like reinforcement learning is not surprisingly convincing. This paper discusses the application of reinforcement techniques in energy systems and how explanations of these models can be helpful.
\end{abstract}


\maketitle




\section{Introduction}
Infrastructure, such as hospitals, power plants, and universities, play an important role in our daily lives. Economic and global development has increased infrastructure complexity. With the help of advancements in technology and automation of labour-intensive tasks, these complex systems require electrical power to be functional and operative. Therefore, to meet the increasing demand for electrical power,  policymakers are encouraged to think about other unconventional sources of energy that not only fulfill the demand but also are more environmentally friendly. Conventional sources of energy, such as oil, gas, and coal energy, are not good for the environment as they emit immense amounts of CO2. Moreover, the reserves of conventional resources are depleting with time, and this scarcity of resources can lead to severe problems for future generations.
Consequently, sustainable distributed energy systems (DERs) are integrated into the energy systems \cite{shengren2023optimal}. While these systems emit less CO2, their inclusion has significantly increased the complexity of the power grids since these DERs are distributed across the power system (potentially far away from consumers) and are weather-dependent. This later property of the DERs adds uncertainty to their nature; hence, it demands the need for proper management and planning of the systems to meet the power demand and supply. 

In 1956 when the term "Artificial intelligence (AI)" was coined \cite{hamet2017artificial}, it was presented as an idea of simulating learning and solving every real-world problem on computers \cite{aiml} with less or no human intervention \cite{hamet2017artificial}. Since computational resources have become readily available, it has become easy for computers to learn the problem-solving process and solve complex tasks in diverse areas with the help of AI and machine learning (ML) algorithms. These techniques have made a foray into every scientific field whether it is healthcare, banking \cite{aiml}, transportation \cite{su14084832}, and many other areas. For instance, in the medical field, researchers have applied AI and ML techniques to diagnose diseases like cancer \cite{huang2020artificial}, outbreak prediction \cite{nayak2018epidemic}, and medical image processing with the help of computer vision \cite{olveres2021new,sathya2020application}. Similarly, researchers have found opportunities to apply AI and ML techniques to address issues of "systems of systems". 

AI has gained significant importance in scientifically addressing infrastructure issues. A lot of studies have been carried out on infrastructure systems and researchers have done an extensive review on how AI can assist us in dealing with problems arising in infrastructure systems like energy, transportation, water systems, and telecommunications \cite{mcmillan2022review}. In energy systems, AI and ML techniques can be used to tackle numerous challenges of the ongoing phase-out of conventional resource usage and the increasing usage of renewable energy. This includes predicting the power generation and load, enhancing monitoring of the system, and finding optimal solutions to reduce power losses \cite{10.1145/3485128}. The above-mentioned problems can be optimally solved with the help of state-of-the-art methods like reinforcement learning. RL methods have proven their credibility in other fields but in energy systems, their application is not convincing which we will discuss in a later section.

Despite the accuracy and precision of the models generated by advanced and state-of-the-art algorithms, policymakers and companies are unable to trust the outputs of many of these algorithms because of their black-box nature \cite{abduljabbar2019applications}. Even the developers of these black-box models are often unable to comprehend how the model processed the different input variables to convert to a certain output function and how the inputs are related to each other \cite{rudin2019we}. Interpretation of these models is crucial in all domains, especially in medicine, energy systems, and other complex infrastructures, as we need to trust the systems concerned with such a vital part of our lives. 

In this paper, we will discuss the applications of ML and RL in energy systems, explain why the interpretation of ML models is important, which techniques are used for explaining these models, and in how existing knowledge should be advanced by new research.

\section{Energy Systems and Reinforcement Learning}
As discussed earlier, energy systems have reached a level of complexity, and with more sensors, an increasing amount of data becomes available so that we can find solutions to these challenges to the energy system with the help of ML-based algorithms. Already, with the digitization of the systems, the usage of machine learning methods has increased drastically. Including DERs made it crucial to monitor the energy systems so they can operate effectively. To model these systems, it is necessary to have a good knowledge of the parameters on which such systems depend, necessitating inference \cite{su14084832}. 

The most commonly used learning types in energy systems are supervised, and unsupervised learning. These techniques are used to forecast the power generation, optimally manage the operational and maintenance cost \cite{shin2021ai}, and schedule the storage and generation operations. Authors in \cite{shin2021ai} have reviewed the number of articles published to apply machine learning methods in the energy domain, and they have found out that after 2012, researchers are more interested in using ML techniques in energy systems, and the number of publications made annually reached up to 2000 \cite{PERERA2021110618}. Authors in \cite{mcmillan2022review} support this increased interest as they reviewed the work of researchers who have used ML-based methods to effectively forecast the generation, manage demand and supply, and monitor the performance of the systems. Hence, both reviews stress the importance of how AI and ML approaches have efficiently solved complex problems in energy systems. 

Even though algorithms behave differently in different domains, authors in \cite{mcmillan2022review} have curated a list of well-performing algorithms that include artificial neural networks (ANN), support vector machines (SVM), decision trees and random forests, and K-mean clustering. Previously, in energy systems, the community leaned mostly toward traditional approaches. However, with the transition and advancements in the methods, researchers in the energy sector are also increasingly adopting ANN-based techniques. Areas like demand forecasting, weather predictions, and price forecasting are popular and are solved by these algorithms \cite{mcmillan2022review}. For example, Hafeez et al  \cite{HAFEEZ2020114915} have proposed a novel electrical load forecasting hybrid model that outperforms previous ANN and LSTM algorithms concerning the accuracy, runtime, and convergence rate. Furthermore, researchers have also reviewed and studied how ANN can be applied to optimize the performance of solar collectors \cite{GHRITLAHRE201875}. They compared the performance with traditional techniques like linear regression and non-linear models and found that ANNs outperformed the other methods.
 
In addition to the use of supervised and unsupervised learning methods, RL is also attracting attention, and researchers are keen to implement RL methods in different domains. Though we have seen RL-based methods applied in different sectors, the researchers of the energy domain seem to be reluctant to adopt RL approaches as can be seen from a low number of RL publications in energy systems \cite{PERERA2021110618}. In RL, the learning agent learns a specific policy based on the feedback given by the environment \cite{kaelbling1996reinforcement}. The RL agent receives a state from state space, and it takes actions either to explore the state space further or to exploit learned rules. This new action leads the agent to a new state after receiving a reward or penalty from the environment \cite{li2017deep}.  This data-driven method helps to learn and create new policies based on trial and error to attain the maximum reward. \revise{RL approaches are more suitable to find optimal solutions and react to dynamic environments than supervised or unsupervised learning. Moreover, in real-time decision-making and optimization problems, the RL-based method can be the best choice \cite{rl}. In recent times, RL methods have proven their worth in different fields like gaming, robotics \cite{rl, 8787888} traffic control systems, advertising, and the stock industry \cite{rl}.} 

Since the complexity of energy systems escalated due to the inclusion of DERs, it is of key importance to optimally control these systems. \revise{Fortunately, the rise of complexity is accompanied by an ever growing amount of operational and real-time data. Processing and harnessing this bulk of data requires state-of-the-art ML and specifically RL approaches to control and optimize the systems \cite{cao2020reinforcement}}. From dispatching generated power optimally over the network to storing energy for later consumption, highly renewable energy systems are more complicated in terms of their system management and therefore offer numerous opportunities for RL applications. The uncertainties of the system like dependency on weather, energy market variations, and improved technologies of the energy systems, state-of-the-art techniques are required to address the problems \cite{PERERA2021110618}. Regardless of the applications and domain, reinforcement learning methodologies have attained success in the scientific field as these methodologies handle uncertainties. Furthermore, for some complex problems, model-free solutions are suitable rather than model-based approaches. Thus, RL methods are a powerful and effective technique for the modern challenges of energy systems \cite{PERERA2021110618}. 

\revise{Concluding, more data becomes available for energy systems, while the integration of DERs increases complexity and requires optimization and control solutions capabale of processing real-time multi-dimensional data efficiently. RL-based approaches are uniquely suitable to address this challenge. The scalability of RL-based approaches has improved with the improvement of state-of-the-art methods like Deep Q network and Proximal Policy Optimization (PPO) since they deal with multi-dimensional data \cite{8787888}. Consequently, we are also proposing to using Deep Q network and PPO to address our research question, as discussed in section 4.}

 \section{Energy Systems and Interpretation of Models}
 Because of the advancements in the field of data science, many data-driven complex tasks are solved with high-level accuracy with the help of ML algorithms. However, most of these models and output functions are uninterpretable, and even the designers of those algorithms are unable to understand how these inputs are converted into the following output. In ML, most of the models are black-box, i.e. these models are generated in an untransparent way that does not reveal how results are obtained, e.g. because a large amount of intrinsically simple units (neural layers) are combined into one extensive complex system. This is in contrast to transparent (or white box) models, such as linear regression, where we clearly see how the model operates but which unfortunately are often performing worse than black boxes \cite{rudin2019we}. AI and ML techniques face backlash from being unreliable and vulnerable to bias. This behaviour of producing unfair and biased results is analyzed by studying AI-designed tools where a famous criminal risk assessment tool was biased in judging African-Americans as criminal, and a beauty pageant winners tool judging black-skinned toned participants unfairly  \cite{8631448, survey}.  Such unreliable results can impact the decision-making process and it will lead to wrong decisions which can later on affect the process and management. Hence, it is of key importance to have interpretability and transparency of the models so that it can help in improving the credibility of models and elevate acceptance of their decisions.

 In this digital age, it becomes inevitable to use AI-based systems for the decision-making processes. The intelligence of systems is helping decision-makers and policymakers to think in different ways\revise{. It is also evident that the fairness, transparency, and interpretability of the predictions or suggestions provided by the AI systems are crucial because these systems have a great impact on human lives \cite{ferreira2021human} }. According to \cite{ferreira2021human}, the most important domains where explainability of these systems are medical and law. However, this idea to make the models explainable is not confined to only these areas. As we discussed earlier, energy systems are also complex systems; hence, the application of AI and ML techniques is beneficial in the decision-making process. However, these ML-based systems are still not fully adopted in reality despite the accuracy and advancement in ML technologies because the black-box nature of such algorithms is a key hurdle \cite{towardsxai}. Furthermore, the unreliability and lack of understanding of these systems is another barrier to adopting ML-based systems specifically in the energy domain \cite{WENNINGER2022118300}. So, to implement these ML-based systems, it is evident that there should be a system that is interpreted by humans and adds to model validation and inspection. Thus, increasing transparency of ML solutions will likely lead to the adaptation of more such models in the energy field. 

 The notion of explanations is defined differently, depending on the context. Sometimes it is presented as an explanation of an intrinsic process which helps users to understand the process \cite{ferreira2021human}, and some researchers categorize the explanations in various sub-categories to achieve different goals like generating trust and transparency, following some renowned regulations, ensuring fairness, generating accountable and reliable models, reduction of misinterpretation in model performance, and validating models \cite{gerlings2020reviewing}. In energy systems, we need explanations as just discussed (e.g. trust, reliability, fairness). Current methods do not yet provide full explanations but so far focus on deriving interpretations of the trained models that identify and quantify the importance of different features or input in generating any model as \cite{ferreira2021human} has done work while modeling energy systems. The necessary explanations also vary depending on the target group, for instance, experts need explanations in terms of finding out the potential significant features in predictions as researchers in \cite{sebastian} have also interpreted their models by computing the feature importance, whereas, for a layperson, the underlying mechanics of the system are not important. Instead, they are interested in actionable and causal explanations of the predictions as mentioned in \cite{WENNINGER2022118300}. 

Though we are witnessing the increasing trend of applying explainable artificial intelligence (XAI) techniques in the energy field, it is still a new and emerging aspect in the field of energy systems, and a lot of work is yet to be done as it is shown in Table~\ref{tab:doc}. \revise{It is important to note that these publications have used a variety of explainability approaches to elaborate a range of energy system problems using again a spectrum of different ML-based methods, including for example SHAP on regression or attention for forecasts \cite{MACHLEV2022100169}}

\begin{table}[]
    \centering
    \begin{tabular}{c|c}
        \hline
        Years & XAI and Energy systems\\
        \hline
        2020-2023  &  64  \\
        2010-2019  &  4  \\
        2000-2009  &  0 
    \end{tabular}
    \caption{This table shows the number of publications published within these years that were related to keywords ``XAI and energy systems``}
    \label{tab:doc}
\end{table}

More and more researchers are shifting their attention to explaining the ML models used in energy systems. There are some areas where the concept of interpreting the models is widely used, including power grid applications, the energy sector, and energy management in buildings \cite{MACHLEV2022100169}. Authors in \cite{MACHLEV2022100169} have listed the commonly used XAI approaches in the energy domain including Shapley additive explanations (SHAP) \cite{9225146,MITRENTSIS2022118473,alova2021machine,9448986},  feature correlation \cite{konstantakopoulos2019design,das2019novel},  and permutation feature importance \cite{en14217367,WENNINGER2022118300}. 

Similar to the explanations of other ML techniques, reinforcement learning is also gaining importance in energy systems so that energy experts can build trust in the systems. \revise{Many researchers have implemented deep reinforcement learning (DRL) to solve complex problems. For example, Huang et al \cite{8787888} have addressed the power energy control system with the help of DRL methods. While in \cite{9347147}, researchers have worked on implementing DRL to address power emergency control, and they have used deep explainer to obtain at least initial interpretations of previously black box models.  However, the explainability of such systems is still mostly missing}.  Thus, a lot of work has yet to be done since interpretations, let alone full explanations, of reinforcement learning algorithms in energy are not very common, as suitable methods are not readily available. To name two examples of RL applications in energy systems that require explanations: \revise{
\begin{itemize}
    \item First, an RL controller of an energy storage system needs explanations because this will help energy experts or laypeople understand and trust the control laws which are learnt by the system. 
    \item Second,  frequency and voltage control of transmission systems could be realized by an RL agent, especially in emergency situations, where humans are too slow to react and existing rule-based systems are too inflexible. 
\end{itemize}
However, actions to stabilize a system have to convince operators so that they take appropriate action and so that potential damage due to failure can be properly investigated. Both of the above-mentioned RL applications make explanations for these models inevitable \cite{MACHLEV2022100169}. }

\section{Our planned contribution}
We wish to focus on explainable reinforcement learning in small-scale energy systems. Due to the stochastic nature of renewable energy sources, it is natural to include energy storage in the energy system to ensure a smoother and more constant supply of power, e.g. to a household. Thermal energy storage (TES) systems are one candidate for balancing the supply and demand of the buildings. Wang et al have already developed a deep Q-network RL approach to optimally control this setup \cite{WANG2023113696}. Utilizing another technology, Elseify et al \cite{ELSEIFY2023108986} considered a scenario with multiple batteries to store photovoltaic (PV) generated power and again obtained an optimal solution using RL methods, specifically an equilibrium optimizer. However, it is crucial to derive explanations for experts and laypeople alike.  Hence, we require suitable methods and proofs-of-concept for explainable reinforcement learning \cite{MACHLEV2022100169}. 

In our research, we want to identify the optimal power management of a building, including balancing renewable generation and consumption via demand response and storage. \revise{Currently, we are considering the use-case of buildings with roof-mounted PV panels that have to balance PV generation, storage and purchases from the power grid. We further assume that these buildings have installed batteries to store excess solar power generated by the roof-mounted PV so that this power can later be used when insufficient power is generated by the PV panels, especially at night and on cloudy days. We search for optimal usage strategies of batteries to minimize the total costs of purchasing power from the connected utility grid. Rather than using supervised and rule-based approaches, we aim to solve this optimization problem with the help of state-of-the-art RL methods like Deep Q-network, and PPO \cite{schulman2017proximal}, so that the agent learns the policies as per the given states of the environment. Technically, we are modeling a single building in terms of demand $D$, grid electricity prices $p$, stored energy in the batteries, and the power generated by the PV panel $G_\text{PV}$. We have used the notion of net demand which is the function of actual demand and power generation of PV panel. \begin{equation}
    \text{Net demand} = f(D, G_\text{PV}) = D - G_\text{PV}.
\end{equation} This net demand indicates the extra power that the building needs from the grid. The agent takes specific actions, for instance, to charge and discharge the battery, and these actions are either rewarded or penalized by the environment. }

\revise{However, we also want to derive useful interpretations and explanations from the model which is a black-box model. Initial steps include interpreting the workings of the model, e.g. by obtaining a ranking of features that have an impact on the predictions with the help of break-down profiles \cite{bdplot}. Thus, we can formulate our research question as \begin{quote}
    Can we use RL to solve an optimization problem in a small-scale energy system (balancing power of a single building) and how can we improve trust in such a black-box model by explaining its learned policies and actions?
\end{quote} In summary, our research contributes in the following way:
\begin{itemize}
    \item Analysing the applicability of RL in solving optimization problems in energy systems, like minimizing the cost of purchasing electricity from the grid by maximizing the consumption of PV panels' generated power
    \item Having a comparative analysis of different state-of-the-art RL methods 
    \item Having explanations of the black-box models to build trust in the outcomes of these models and analysing how features impact the predictions
\end{itemize}
Later on, we will look for full explanations of why the model took a certain action and why it did not take a different one (counterfactual explanations), and ideally, generate these explanations dynamically to explain the rationale of each step and every point in time.} The goal is to increase trust and transparency so that these systems can be implemented at large scales and experts and laypeople feel confident while adapting such systems. 

\section{Conclusion}
Energy systems, as critical infrastructure and with many intrinsic optimization problems, are excellent systems to apply explainable ML and RL techniques. Unfortunately, XAI is still used too reluctantly for supervised problems and even more rarely for RL. Hence, we propose to bridge this gap and develop suitable explainable reinforcement learning methods for energy systems.

\section{Acknowledgements}
We gratefully acknowledge funding from the Helmholtz Association under grant no. VH-NG-1727 and the Networking Fund through Helmholtz AI.







\bibliographystyle{IEEEtran}
\bibliography{references}
\end{document}